\def\paperID{8582}
\def\confName{CVPR}
\def\confYear{2024}

\newif\ifreview 
\newif\ifarxiv \newcommand{\arxiv}{\arxivtrue}
\newif\ifcamera 
\newif\ifrebuttal 

\arxiv 
\pdfoutput=1
\documentclass[10pt,twocolumn,letterpaper]{article}

\newcommand{\eg}{\textit{e.g.}}

\newcommand{\workname}{FlowVid\xspace}
 
\renewcommand{\thefootnote}{\fnsymbol{footnote}}

\ifreview \usepackage[review]{cvpr} \fi
\ifarxiv \usepackage[pagenumbers]{cvpr} \fi
\ifrebuttal \usepackage[rebuttal]{cvpr} \fi
\ifcamera \usepackage{cvpr} \fi

\usepackage{graphicx}	
\usepackage{amsmath}	
\usepackage{amssymb}	
\usepackage{booktabs}
\usepackage{times}
\usepackage{microtype}
\usepackage{epsfig}
\usepackage[table,xcdraw,dvipsnames]{xcolor}
\usepackage{caption}
\usepackage{float}
\usepackage{placeins}
\usepackage{color, colortbl}
\usepackage{stfloats}
\usepackage{enumitem}
\usepackage{tabularx}
\usepackage{xstring}
\usepackage{multirow}
\usepackage{xspace}
\usepackage{url}
\usepackage{subcaption}
\usepackage{xcolor}
\usepackage[hang,flushmargin]{footmisc}
\ifcamera \usepackage[accsupp]{axessibility} \fi

% ...

\ifarxiv  \fi

\newcommand{\R}[1]{{%
    \textbf{%
        \ifstrequal{#1}{1}{\textcolor{red}{R#1}}{%
        \ifstrequal{#1}{2}{\textcolor{blue}{R#1}}{%
        \ifstrequal{#1}{3}{\textcolor{magenta}{R#1}}{%
        \ifstrequal{#1}{4}{\textcolor{teal}{R#1}}{%
                           \textcolor{cyan}{R#1}%
        }}}}%
    }%
}}

\usepackage{xr-hyper}

\makeatletter
\newcommand*{\addFileDependency}[1]{
  \typeout{(#1)}
  \@addtofilelist{#1}
  \IfFileExists{#1}{}{\typeout{No file #1.}}
}

\makeatother

\definecolor{cvprblue}{rgb}{0.21,0.49,0.74}
\usepackage[pagebackref,breaklinks,colorlinks,citecolor=cvprblue]{hyperref}
\usepackage[capitalize]{cleveref}
\usepackage{multicol}
\usepackage{fancyhdr}

\crefname{section}{Sec.}{Secs.}
\crefname{table}{Table}{Tables}
\crefname{figure}{Fig.}{Figs.}

\begin{document}

\twocolumn[{

\title{\workname: Taming Imperfect Optical Flows 
for Consistent Video-to-Video Synthesis \vspace{-2em}
}
\author{Feng Liang\footnotemark ~~\textsuperscript{\rm 1}, 
        Bichen Wu\footnotemark ~~\textsuperscript{\rm 2}, 
        Jialiang Wang\textsuperscript{\rm 2},
        Licheng Yu\textsuperscript{\rm 2},
        Kunpeng Li\textsuperscript{\rm 2}, 
        Yinan Zhao\textsuperscript{\rm 2}, 
        Ishan Misra\textsuperscript{\rm 2}, \\
        Jia-Bin Huang\textsuperscript{\rm 2},
        Peizhao Zhang\textsuperscript{\rm 2}, 
        Peter Vajda\textsuperscript{\rm 2}, 
        Diana Marculescu\textsuperscript{\rm 1}\\
\textsuperscript{\rm 1}The University of Texas at Austin, \textsuperscript{\rm 2}Meta GenAI\\
\texttt{\{jeffliang,dianam\}@utexas.edu}, \texttt{\{wbc,stzpz,vajdap\}@meta.com}\\
\texttt{\href{https://jeff-liangf.github.io/projects/flowvid}{https://jeff-liangf.github.io/projects/flowvid}}
}
\maketitle

\vspace{-2.4em}

\begin{center}
    \centering
    \includegraphics[width=0.98\textwidth]{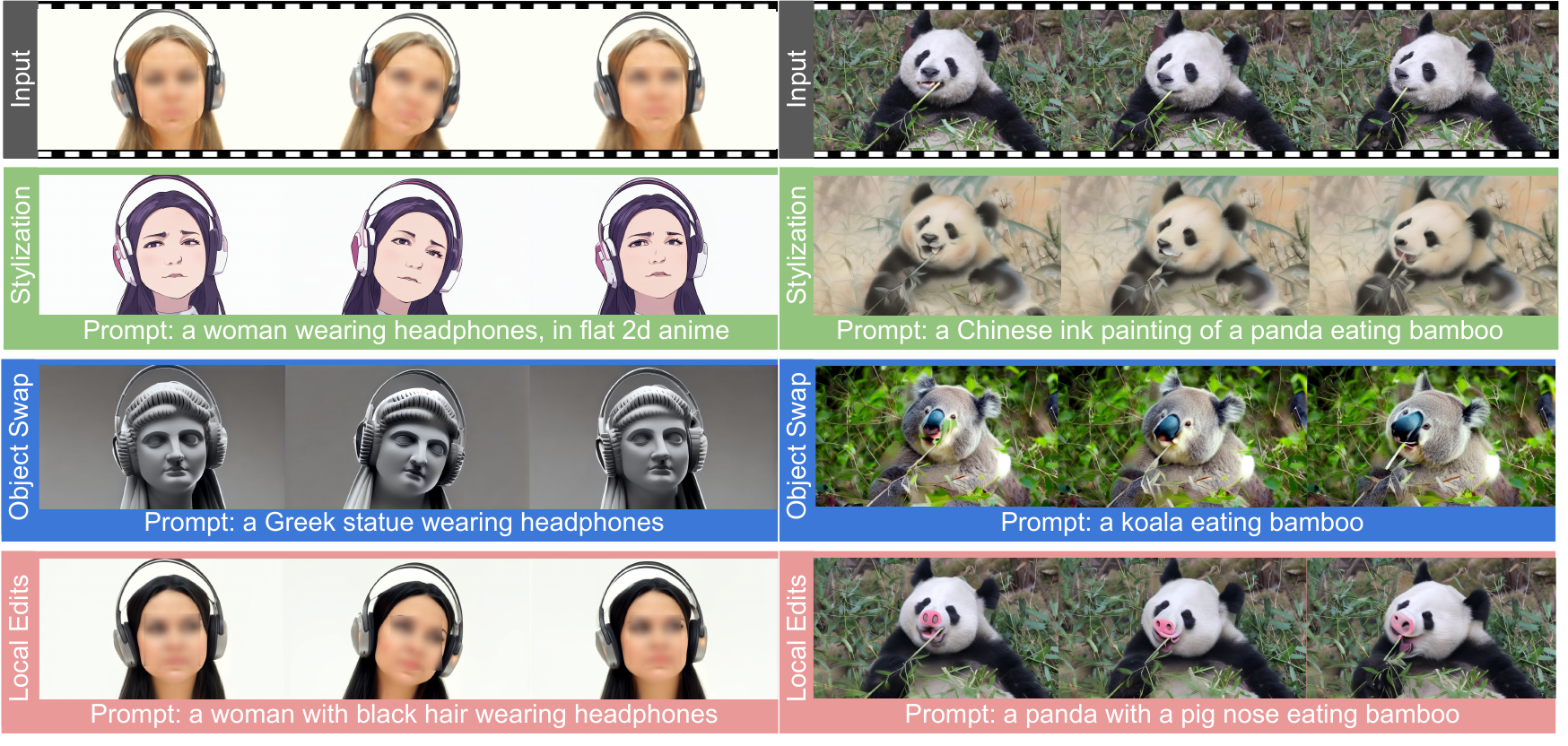}
    \vspace{-0.5em}
    \captionof{figure}{
    We present \textbf{FlowVid} to synthesize a consistent video given an input video and a target prompt. Our model supports multiple applications: (1) global stylization, such as converting the video to 2D anime (2) object swap, such as turning the panda into a koala bear (3) local edit, such as adding a pig nose to a panda.}
    \label{fig:main_figure}
\end{center}
}]

\footnotetext{\textsuperscript{*}Work partially done during an internship at Meta GenAI.}
\footnotetext{\textsuperscript{†}Corresponding author.}

\renewcommand{\thefootnote}{\arabic{footnote}}

\begin{abstract}
Diffusion models have transformed the image-to-image (I2I) synthesis and are now permeating into videos. However, the advancement of video-to-video (V2V) synthesis has been hampered by the challenge of maintaining temporal consistency across video frames. This paper proposes a consistent V2V synthesis framework by jointly leveraging spatial conditions and temporal optical flow clues within the source video. Contrary to prior methods that strictly adhere to optical flow, our approach harnesses its benefits while handling the imperfection in flow estimation. We encode the optical flow via warping from the first frame and serve it as a supplementary reference in the diffusion model. This enables our model for video synthesis by editing the first frame with any prevalent I2I models and then propagating edits to successive frames. Our V2V model, \workname, demonstrates remarkable properties: (1) Flexibility: \workname works seamlessly with existing I2I models, facilitating various modifications, including stylization, object swaps, and local edits. (2) Efficiency: Generation of a 4-second video with 30 FPS and 512$\times$512 resolution takes only 1.5 minutes, which is 3.1$\times$, 7.2$\times$, and 10.5$\times$ faster than CoDeF, Rerender, and TokenFlow, respectively. (3) High-quality: In user studies, our \workname is preferred 45.7\% of the time, outperforming CoDeF (3.5\%), Rerender (10.2\%), and TokenFlow (40.4\%).
\end{abstract}

\begin{figure}[t]
    \centering
	\includegraphics[width=1.0\columnwidth]{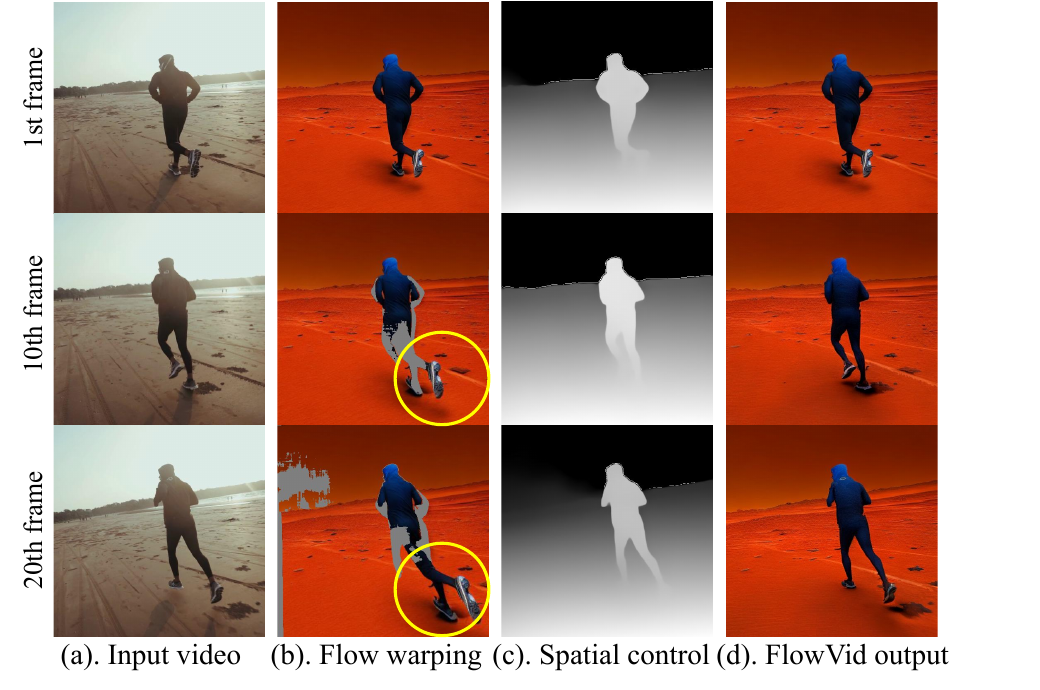}
	\caption{(a) Input video: \texttt{'a man is running on beach'}. (b) We edit the 1st frame with \texttt{'a man is running on Mars'}, then conduct flow warping from the 1st frame to the 10th and 20th frames (using input video flow). Flow estimation of legs is inaccurate. (c) Our \workname uses spatial controls to rectify the inaccurate flow. (d) Our consistent video synthesis results.}
 \vspace{-1em}
	\label{fig:teaser}
\end{figure}


\section{Introduction}
\label{sec:intro}

Text-guided Video-to-video (V2V) synthesis, which aims to modify the input video according to given text prompts, has wide applications in various domains, such as short-video creation and more broadly in the film industry. 
Notable advancements have been seen in text-guided Image-to-Image (I2I) synthesis~\cite{ip2p_brooks2023instructpix2pix, p2p_hertz2022prompt, pnp_tumanyan2023plug, t2iadapter_mou2023t2i}, greatly supported by large pre-trained text-to-image diffusion models~\cite{dalle2_ramesh2022hierarchical, ldm_rombach2022high, imagen_saharia2022photorealistic}.
However, V2V synthesis remains a formidable task.
In contrast to still images, videos encompass an added temporal dimension.
Due to the ambiguity of text, there are countless ways to edit frames so they align with the target prompt.
Consequently, naively applying I2I models on videos often produces unsatisfactory pixel flickering between frames.

To improve frame consistency, pioneering studies edit multiple frames jointly by inflating the image model with spatial-temporal attention~\cite{tuneavideo_wu2023tune, text2videozero_khachatryan2023text2video, ceylan2023pix2video, fatezero_qi2023fatezero}.
While these methods offer improvements, they do not fully attain the sought-after temporal consistency. 
This is because the motion within videos is merely retained in an \emph{implicit} manner within the attention module.
Furthermore, a growing body of research employs \emph{explicit} optical flow guidance from videos.
Specifically, flow is used to derive pixel correspondence, resulting in a pixel-wise mapping between two frames.
The correspondence is later utilized to obtain occlusion masks for inpainting~\cite{rerender_yang2023rerender,hu2023videocontrolnet} or to construct a canonical image~\cite{codef_ouyang2023codef}
However, these hard constraints can be problematic if flow estimation is inaccurate, which is often observed when the flow is determined through a pre-trained model~\cite{teed2020raft, xu2022gmflow, unimatch_xu2023unifying}.

In this paper, we propose to harness the benefits of optical flow while handling the imperfection in flow estimation.
Specifically, we perform flow warping from the first frame to subsequent frames. These warped frames are expected to follow the structure of the original frames but contain some occluded regions (marked as gray), as shown in Figure~\ref{fig:teaser}(b). 
If we use flow as hard constraints, such as inpainting~\cite{rerender_yang2023rerender,hu2023videocontrolnet} the occluded regions, the inaccurate legs estimation would persist, leading to an undesirable outcome.
We seek to include an additional spatial condition, such as a depth map in Figure~\ref{fig:teaser}(c), along with a temporal flow condition.
The legs' position is correct in spatial conditions, and therefore, the joint spatial-temporal condition would rectify the imperfect optical flow, resulting in consistent results in Figure~\ref{fig:teaser}(d).

We build a video diffusion model upon an inflated spatial controlled I2I model.
We train the model to predict the input video using spatial conditions (\eg, depth maps) and temporal conditions (flow-warped video). During generation, we employ an \emph{edit-propagate} procedure: (1) Edit the first frame with prevalent I2I models. (2) Propagate the edits throughout the video using our trained model.
The decoupled design allows us to adopt an autoregressive mechanism: the current batch's last frame can be the next batch's first frame, allowing us to generate lengthy videos.

We train our model with 100k real videos from ShutterStock~\cite{ShutterstockVideo}, and it generalizes well to different types of modifications, such as stylization, object swaps, and local edits, as seen in Figure~\ref{fig:main_figure}.
Compared with existing V2V methods, our \workname demonstrates significant advantages in terms of efficiency and quality.
Our \workname can generate 120 frames (4 seconds at 30 FPS) in high-resolution (512$\times$512) in just 1.5 minutes on one A-100 GPU, which is  3.1$\times$, 7.2$\times$ and 10.5$\times$ faster than state-of-the-art methods CoDeF~\cite{codef_ouyang2023codef} (4.6 minutes) Rerender~\cite{rerender_yang2023rerender} (10.8 minutes), and TokenFlow~\cite{geyer2023tokenflow} (15.8 minutes).
We conducted a user study on 25 DAVIS~\cite{davis_pont20172017} videos and designed 115 prompts. Results show that our method is more robust and achieves a preference rate of 45.7\% compared to CoDeF (3.5\%) Rerender (10.2\%) and TokenFlow (40.4\%)

Our contributions are summarized as follows: (1) We introduce \workname, a V2V synthesis method that harnesses the benefits of optical flow, while delicately handling the imperfection in flow estimation.
(2) Our decoupled edit-propagate design supports multiple applications, including stylization, object swap, and local editing. 
Furthermore, it empowers us to generate lengthy videos via autoregressive evaluation. (3) Large-scale human evaluation indicates the efficiency and high generation quality of \workname. 
\section{Related Work}
\label{sec:related}

\subsection{Image-to-image Diffusion Models}

Benefiting from large-scale pre-trained text-to-image (T2I) diffusion models~\cite{ldm_rombach2022high, imagen_saharia2022photorealistic, balaji2022ediffi, dai2023emu}, progress has been made in text-based image-to-image (I2I) generation~\cite{ t2iadapter_mou2023t2i,pnp_tumanyan2023plug,kawar2023imagic,p2p_hertz2022prompt,zhang2023sine,parmar2023zero, meng2021sdedit, couairon2022diffedit}.
Beginning with image editing methods, Prompt-to-prompt~\cite{p2p_hertz2022prompt} and PNP~\cite{pnp_tumanyan2023plug} manipulate the attentions in the diffusion process to edit images according to target prompts. Instruct-pix2pix~\cite{ip2p_brooks2023instructpix2pix} goes a step further by training an I2I model that can directly interpret and follow human instructions.
More recently, I2I methods have extended user control by allowing the inclusion of reference images to precisely define target image compositions.
Notably, ControlNet, T2I-Adapter~\cite{t2iadapter_mou2023t2i}, and Composer~\cite{huang2023composer} have introduced spatial conditions, such as depth maps, enabling generated images to replicate the structure of the reference.
Our method falls into this category as we aim to generate a new video while incorporating the spatial composition in the original one.
However, it's important to note that simply applying these I2I methods to individual video frames can yield unsatisfactory results due to the inherent challenge of maintaining consistency across independently generated frames (per-frame results can be found in Section~\ref{exp:qualitative_results}).

\subsection{Video-to-video Diffusion Models}
To jointly generate coherent multiple frames, it is now a common standard to inflate image models to video: replacing spatial-only attention with spatial-temporal attention.
For instance, Tune-A-Video \cite{tuneavideo_wu2023tune}, Vid-to-vid zero~\cite{vid2vidzero_wang2023zero}, Text2video-zero \cite{text2videozero_khachatryan2023text2video}, Pix2Video~\cite{ceylan2023pix2video} and FateZero~\cite{fatezero_qi2023fatezero} performs cross-frame attention of each frame on anchor frame, usually the first frame and the previous frame to preserve appearance consistency. 
TokenFlow \cite{geyer2023tokenflow} further explicitly enforces semantic correspondences of diffusion features across frames to improve consistency.
Furthermore, more works are adding spatial controls, \eg, depth map to constraint the generation. 
Zhang's ControlVideo~\cite{zhang2023controlvideo} proposes to extend image-based ControlNet to the video domain with full cross-frame attention. 
Gen-1~\cite{gen1_esser2023structure}, VideoComposer~\cite{wang2023videocomposer}, Control-A-Video~\cite{controlavideo_chen2023control} and Zhao's ControlVideo~\cite{zhao2023controlvideo} train V2V models with paired spatial controls and video data.
Our method falls in the same category but it also includes the imperfect temporal flow information into the training process alongside spatial controls. This addition enhances the overall robustness and adaptability of our method.

Another line of work is representing video as 2D images, as seen in methods like layered atlas~\cite{kasten2021layered}, Text2Live~\cite{bar2022text2live}, shape-aware-edit~\cite{lee2023shape}, and CoDeF~\cite{codef_ouyang2023codef}. However, these methods often require per-video optimization and they also face performance degradation when dealing with large motion, which challenges the creation of image representations.

\subsection{Optical flow for video-to-video synthesis}
The use of optical flow to propagate edits across frames has been explored even before the advent of diffusion models, as demonstrated by the well-known Ebsythn~\cite{jamrivska2019stylizing} approach.
In the era of diffusion models, 
Chu's Video ControlNet~\cite{chu2023video} employs the ground-truth (gt) optical flow from synthetic videos to enforce temporal consistency among corresponding pixels across frames. However, it's important to note that ground-truth flow is typically unavailable in real-world videos, where flow is commonly estimated using pretrained models~\cite{teed2020raft, xu2022gmflow, unimatch_xu2023unifying}.
Recent methods like Rerender~\cite{rerender_yang2023rerender}, MeDM~\cite{chu2023medm}, and Hu's VideoControlNet~\cite{hu2023videocontrolnet} use estimated flow to generate occlusion masks for in-painting. In other words, these methods "force" the overlapped regions to remain consistent based on flow estimates. 
Similarly, CoDeF~\cite{codef_ouyang2023codef} utilizes flow to guide the generation of canonical images.
These approaches all assume that flow could be treated as an accurate supervision signal that must be strictly adhered to.
In contrast, our \workname recognizes the imperfections inherent in flow estimation and presents an approach that leverages its potential without imposing rigid constraints.

\section{Preliminary}
\label{sec:method}

\begin{figure*}[t]
    \centering
	\includegraphics[width=2.0\columnwidth]{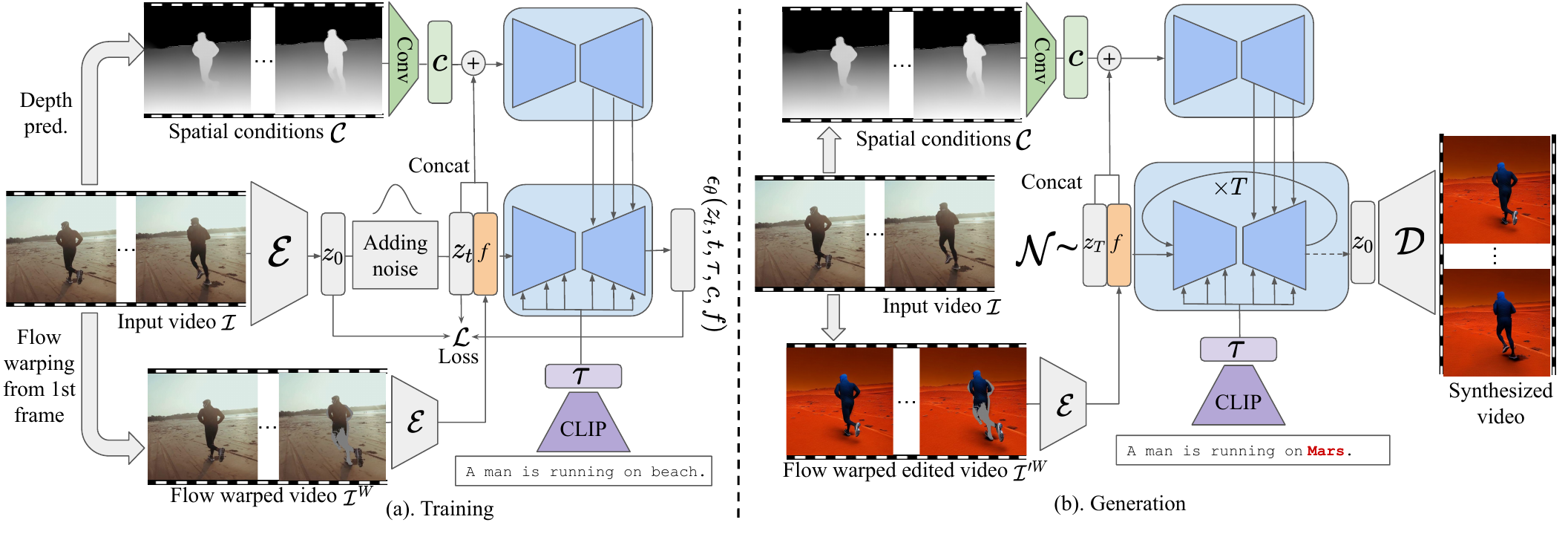}
	\caption{\textbf{Overview of our \workname.} (a) Training: we first get the spatial conditions (predicted depth maps) and estimated optical flow from the input video. For all frames, we use flow to perform warping from the first frame. The resulting flow-warped video is expected to have a similar structure as the input video but with some occluded regions (marked as gray, better zoomed in). We train a video diffusion model with spatial conditions $c$ and flow information $f$. (b) Generation: we edit the first frame with existing I2I models and use the flow in the input video to get the flow warped edited video. The flow condition spatial condition jointly guides the output video synthesis.
	}
	\label{fig:method}
\end{figure*}


\paragraph{Latent Diffusion Models} 
Denoising Diffusion Probabilistic Models (DDPM)~\cite{ddpm_ho2020denoising} generate images through a progressive noise removal process applied to an initial Gaussian noise, carried out for $T$ time steps.
Latent Diffusion models~\cite{ldm_rombach2022high} conduct diffusion process in latent space to make it more efficient.
Specifically, an encoder $\mathcal{E}$ compresses an image $I \in \mathbb{R}^{H \times W \times 3}$ to a low-resolution latent code $z=\mathcal{E}(I) \in \mathbb{R}^{H/8 \times W/8 \times 4}$. 
Given $z_0 := z$, the Gaussian noise is gradually added on $z_0$ with time step $t$ to get noisy sample $z_t$.
Text prompt $\tau$ is also a commonly used condition.
A time-conditional U-Net $\epsilon_\theta$ is trained to reverse the process with the loss function: 

\vspace{-0.8em}
\begin{equation}
    \mathcal{L}_{LDM} =  \mathbb{E}_{z_0, t, \tau, \epsilon \sim \mathcal{N}(0, 1)} \| \epsilon - \epsilon_\theta (z_t, t, \tau) \|_2^2
\label{eq:ldm}
\vspace{-1em}
\end{equation}

\paragraph{ControlNet} 
ControlNet provides additional spatial conditions, such as canny edge~\cite{canny1986computational} and depth map~\cite{midas_ranftl2020towards},  to control the generation of images. More specifically, spatial conditions $C \in \mathbb{R}^{H \times W \times 3}$ are first converted to latents $c \in \mathbb{R}^{H/8 \times W/8 \times 4}$ via several learnable convolutional layers. Spatial latent $c$, added by input latent $z_t$, is passed to a copy of the pre-trained diffusion model, more known as ControlNet. The ControlNet interacts with the diffusion model in multiple feature resolutions to add spatial guidance during image generation.
ControlNet rewrites Equation~\ref{eq:ldm} to 

\vspace{-1em}
\begin{equation}
    \mathcal{L}_{CN} =  \mathbb{E}_{z_0, t, \tau, c, \epsilon \sim \mathcal{N}(0, 1)} \| \epsilon - \epsilon_\theta (z_t, t, \tau, c) \|_2^2
\label{eq:controlnet}
\end{equation}

\section{FlowVid}
For video-to-video generation, given an input video with $N$ frames $\mathcal{I} = \{I_1, \dots, I_N\}$ and a text prompt $\tau$, the goal is transfer it to a new video $\mathcal{I'} = \{I'_1, \dots, I'_N\}$ which adheres to the provided prompt $\tau'$, while keeping consistency across frame.
We first discuss how we inflate the image-to-image diffusion model, such as ControlNet to video, with spatial-temporal attention~\cite{tuneavideo_wu2023tune, text2videozero_khachatryan2023text2video, ceylan2023pix2video, fatezero_qi2023fatezero} (Section~\ref{method:inflate_st_attn})
Then, we introduce how to incorporate imperfect optical flow as a condition into our model (Section~\ref{method:training}). 
Lastly, we introduce the edit-propagate design for generation (Section~\ref{method:inference}).

\subsection{Inflating image U-Net to accommodate video}
\label{method:inflate_st_attn}

The latent diffusion models (LDMs) are built upon the architecture of U-Net, which comprises multiple encoder and decoder blocks. 
Each block has two components: a residual convolutional module and a transformer module. The transformer module, in particular, comprises a spatial self-attention layer, a cross-attention layer, and a feed-forward network.
To extend the U-Net architecture to accommodate an additional temporal dimension, we first modify all the 2D layers within the convolutional module to pseudo-3D layers and add an extra temporal self-attention layer~\cite{vdm_ho2022video}.
Following common practice~\cite{vdm_ho2022video,tuneavideo_wu2023tune, text2videozero_khachatryan2023text2video, ceylan2023pix2video, fatezero_qi2023fatezero}, we further adapt the spatial self-attention layer to a spatial-temporal self-attention layer. For video frame $I_i$, the attention matrix would take the information from the first frame $I_1$ and the previous frame $I_{i-1}$. Specifically, we obtain the query feature from frame $I_i$, while getting the key and value features from $I_1$ and $I_{i-1}$. The $\mathrm{Attention}(Q,K,V)$ of spatial-temporal self-attention could be written as 

\begin{equation}
\footnotesize{
Q=W^Q z_{I_i}, K=W^K \left[ z_{I_1}, z_{I_{i-1}} \right], V=W^V \left[ z_{I_1}, z_{I_{i-1}} \right]}
\label{eq:st_attention}
\end{equation}
where $W^Q$, $W^K$, and $W^V$ are learnable matrices that project the inputs to query, key, and value. $z_{I_i}$ is the latent for frame $I_i$. $\left[ \cdot \right]$ denotes concatenation operation. Our model includes an additional ControlNet U-Net that processes spatial conditions. 
We discovered that it suffices only to expand the major U-Net, as the output from the ControlNet U-Net is integrated into this major U-Net.

\subsection{Training with joint spatial-temporal conditions}
\label{method:training}

Upon expanding the image model, a straightforward method might be to train the video model using paired depth-video data. Yet, our empirical analysis indicates that this leads to sub-optimal results, as detailed in the ablation study in Section~\ref{sec:ablation_conditions}. We hypothesize that this method neglects the temporal clue within the video, making the frame consistency hard to maintain.
While some studies, such as Rerender~\cite{rerender_yang2023rerender} and CoDeF~\cite{codef_ouyang2023codef}, incorporate optical flow in video synthesis, they typically apply it as a rigid constraint. In contrast, our approach uses flow as a soft condition, allowing us to manage the imperfections commonly found in flow estimation.

Given a sequence of frames $\mathcal{I}$, we calculate the flow between the first frame $I_1$ and other frames $I_i$, using a pre-trained flow estimation model UniMatch~\cite{unimatch_xu2023unifying}. 
We denote the $\mathcal{F}_{1 \rightarrow i}$ and $\mathcal{F}_{i \rightarrow 1}$ as the forward and backward flow. 
Using forward-backward consistency check~\cite{meister2018unflow}, we can derive forward and backward occlusion masks $O^{fwd}_{1 \rightarrow i}$ and $O^{bwd}_{i \rightarrow 1}$.
Use backward flow $\mathcal{F}_{i \rightarrow 1}$ and occlusion $O^{bwd}_{i \rightarrow 1}$, we can perform $Warp$ operation over the first frame $I_1$ to get $I^W_i$. Intuitively, warped $i^{th}$ frame $I^W_i$ has the same layout as the original frame $I_i$ but the pixels are from the first frame $I_1$.
Due to occlusion, some blank areas could be in the $I^W_i$ (marked as gray in Figure~\ref{fig:method}). 

We denote the sequence of warped frames as flow warped video $\mathcal{I}^W = \{I^W_1, \dots, I^W_N\}$. We feed $\mathcal{I}^W$ into the same encoder $\mathcal{E}$ to convert it into a latent representation $f$. This latent representation is then concatenated with the noisy input $z_t$ to serve as conditions. To handle the increased channel dimensions of $f$, we augment the first layer of the U-Net with additional channels, initializing these new channels with zero weights. 
We also integrate this concatenated flow information into the spatial ControlNet U-Net, reconfiguring its initial layer to include additional channels. With this introduced flow information $f$, we modify Equation~\ref{eq:controlnet} as: 
\begin{equation}
    \mathcal{L}_{FlowVid} =  \mathbb{E}_{z_0, t, \tau, c, f, \epsilon \sim \mathcal{N}(0, 1)} \| \epsilon - \epsilon_\theta (z_t, t, \tau, c, f) \|_2^2
\label{eq:flowvid}
\end{equation}

Throughout the development of our experiments, two particular design choices have been proven crucial for enhancing our final results. First, we opted for $v$-parameterization~\cite{vpredicton_salimans2022progressive}, rather than the more commonly used $\epsilon$-parameterization. This finding is consistent with other video diffusion models, such as Gen-1~\cite{gen1_esser2023structure} and Imagen Video~\cite{imagenvideo_ho2022imagen} (see ablation in Section~\ref{sec:v_epsilon}). Second, incorporating additional elements beyond the flow-warped video would further improve the performance. Specifically, including the first frame as a constant video sequence, $\mathcal{I}^{1st} = \{I_1, \dots, I_1\}$, and integrating the occlusion masks $\mathcal{O} = \{O^{bwd}_{1 \rightarrow 1}, \dots, O^{bwd}_{N \rightarrow 1}\}$ enhanced the overall output quality. We process $\mathcal{I}^{1st}$ by transforming it into a latent representation and then concatenating it with the noisy latent, similar to processing $\mathcal{I}^{W}$. For $\mathcal{O}$, we resize the binary mask to match the latent size before concatenating it with the noisy latent. 
Further study is included in Section~\ref{sec:ablation_conditions}.

\subsection{Generation: edit the first frame then propagate}
\label{method:inference}

During the generation, we want to transfer the input video $\mathcal{I}$ to a new video $\mathcal{I'}$ with the target prompt $\tau'$.
To effectively leverage the prevalent I2I models, we adopt an edit-propagate method. 
This begins with editing the first frame $I_1$ using I2I models, resulting in an edited first frame $I'_1$. 
We then propagate the edits to subsequent $i^{th}$ frame by using the flow $\mathcal{F}_{i \rightarrow 1}$ and the occlusion mask $O^{bwd}_{i \rightarrow 1}$, derived from the input video $\mathcal{I}$. 
This process yields the flow-warped edited video $\mathcal{I'}^W = \{I'^W_1, \dots, I'^W_N\}$.
We input $\mathcal{I'}^W$ into the same encoder $\mathcal{E}$ and concatenate the resulting flow latent $f$ with a randomly initialized Gaussian noise $z_T$ drawn from the normal distribution $\mathcal{N}$. 
The spatial conditions from the input video are also used to guide the structural layout of the synthesized video.
Intuitively, the flow-warped edited video serves as a texture reference while spatial controls regularize the generation, especially when we have inaccurate flow. 
After DDIM denoising, the denoised latent $z_0$ is brought back to pixel space with a decoder $\mathcal{D}$ to get the final output.

\begin{figure}[t]
    \centering
	\includegraphics[width=1.0\columnwidth]{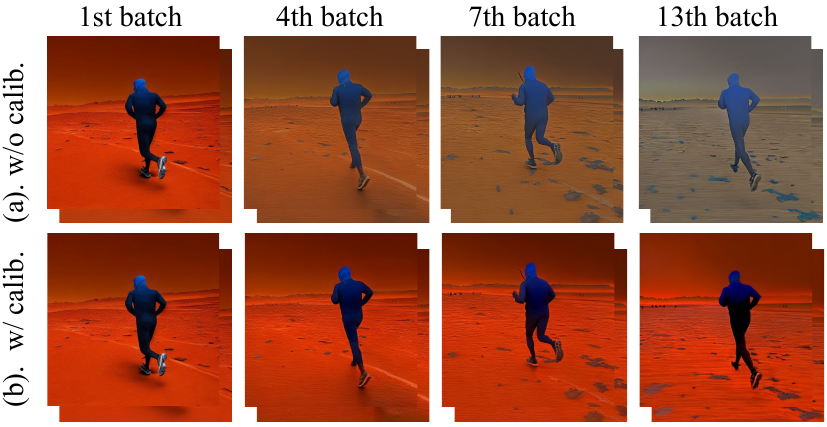}
	\caption{\textbf{Effect of color calibration in autoregressive evaluation.} (a) When the autoregressive evaluation goes from the 1st batch to the 13th batch, the results without color calibration become gray. (b) The results are more stable with the proposed color calibration.}
	\label{fig:ar_color_calib}
\end{figure}

In addition to offering the flexibility to select I2I models for initial frame edits, our model is inherently capable of producing extended video clips in an autoregressive manner. 
Once the first $N$ edited frames $\{I'_1, \dots, I'_N\}$ are generated, the $N^{th}$ frame $I'_N$ can be used as the starting point for editing the subsequent batch of frames $\{I_N, \dots, I_{2N-1}\}$. However, a straightforward autoregressive approach may lead to a \textit{grayish} effect, where the generated images progressively become grayer, see Figure~\ref{fig:ar_color_calib}(a).
We believe this is a consequence of the lossy nature of the encoder and decoder, a phenomenon also noted in Rerender~\cite{rerender_yang2023rerender}.
To mitigate this issue, we introduce a simple global color calibration technique that effectively reduces the graying effect. Specifically, for each frame $I'_j$ in the generated sequence $\{I'_1, \dots, I'_{M(N-1)+1}\}$, where $M$ is the number of autoregressive batches, we calibrate its mean and variance to match those of $I'_1$. The effect of calibration is shown in Figure~\ref{fig:ar_color_calib}(b), where the global color is preserved across autoregressive batches.

\vspace{-0.5em}
\begin{equation}
    I''_j = \left( \frac{I'_j - \text{mean}(I'_j)}{\text{std}(I'_j)} \right) \times \text{std}(I'_1) + \text{mean}(I'_1)
\label{eq:color_calib}
\end{equation}

Another advantageous strategy we discovered is the integration of self-attention features from DDIM inversion, a technique also employed in works like FateZero~\cite{fatezero_qi2023fatezero} and TokenFlow~\cite{geyer2023tokenflow}. This integration helps preserve the original structure and motion in the input video. 
Concretely, we use DDIM inversion to invert the input video with the original prompt and save the intermediate self-attention maps at various timesteps, usually 20. 
During the generation with the target prompt, we substitute the keys and values in the self-attention modules with these pre-stored maps.
Then, during the generation process guided by the target prompt, we replace the keys and values within the self-attention modules with previously saved corresponding maps. 

\section{Experiments}
\label{sec:experiments}

\begin{figure*}[t]
    \centering
	\includegraphics[width=2.0\columnwidth]{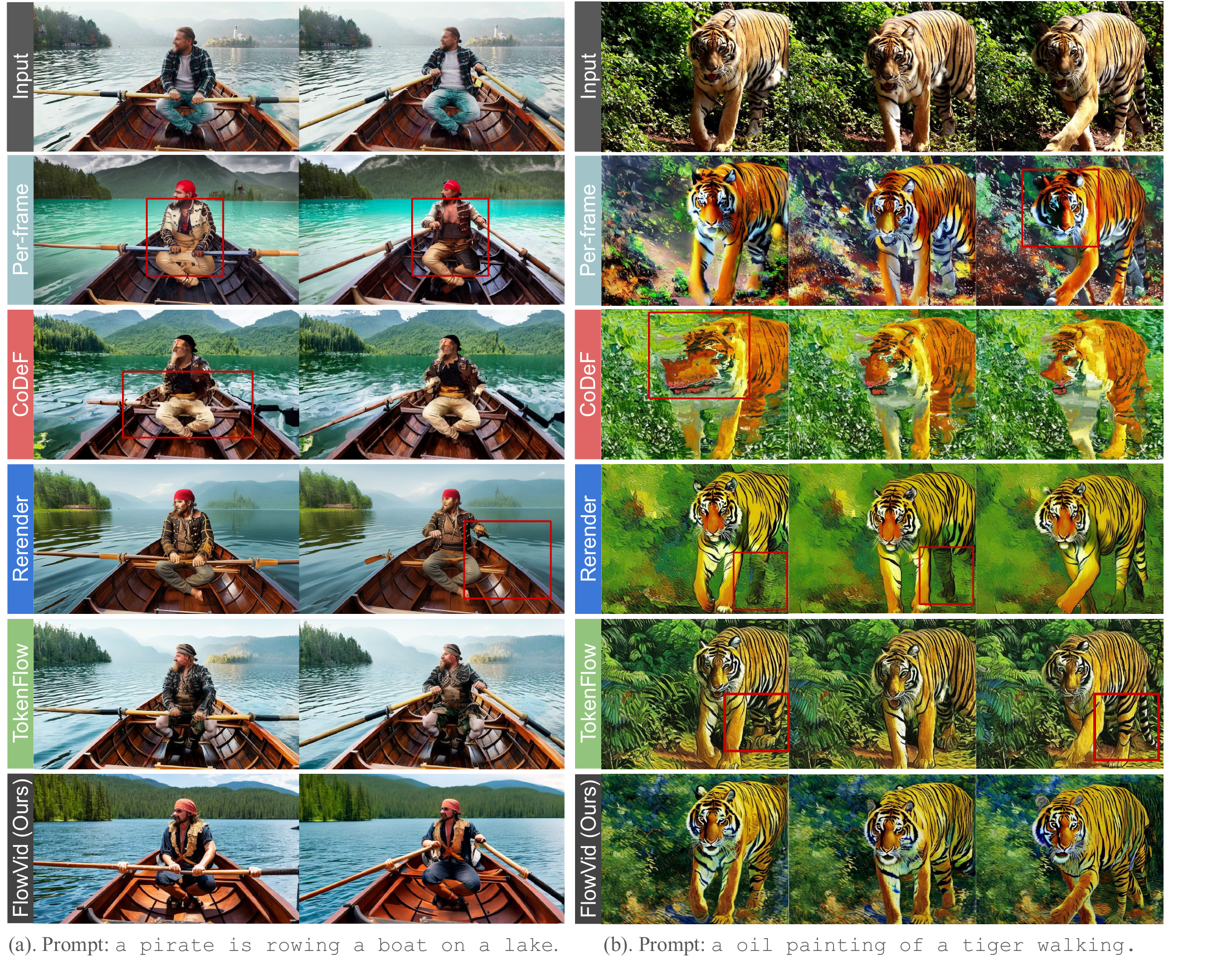}
    \vspace{-0.5em}
	\caption{\textbf{Qualitative comparison with representative V2V models.} Our method stands out in terms of prompt alignment and overall video quality. We highly encourage readers to refer to video comparisons in our supplementary videos.
	}
	\vspace{-0.8em}
	\label{fig:main_comparsion}
\end{figure*}

\subsection{Settings}
\label{exp:settings}
\paragraph{Implementation Details}
We train our model with 100k videos from Shutterstock~\cite{ShutterstockVideo}. For each training video, we sequentially sample 16 frames with interval \{2,4,8\}, which represent videos lasting \{1,2,4\} seconds (taking videos with FPS of 30). The resolution of all images, including input frames, spatial condition images, and flow warped frames, is set to 512$\times$512 via center crop.
We train the model with a batch size of 1 per GPU and a total batch size of 8 with 8 GPUs. We employ AdamW optimizer \cite{adamw_loshchilov2017decoupled} with a learning rate of 1e-5 for 100k iterations.
As detailed in our method, we train the major U-Net and ControlNet U-Net joint branches with $v$-parameterization~\cite{vpredicton_salimans2022progressive}.
The training takes four days on one 8-A100-80G node.

During generation, we first generate keyframes with our trained model and then use an off-the-shelf frame interpolation model, such as RIFE~\cite{huang2022rife}, to generate non-key frames. 
By default, we produce 16 key frames at an interval of 4, corresponding to a 2-second clip at 8 FPS. 
Then, we use RIFE to interpolate the results to 32 FPS.
We employ classifier-free guidance~\cite{cfg_ho2022classifier} with a scale of 7.5 and use 20 inference sampling steps. Additionally, the Zero SNR noise scheduler~\cite{zerosnr_lin2023common} is utilized. We also fuse the self-attention features obtained during the DDIM inversion of corresponding key frames from the input video, following FateZero~\cite{fatezero_qi2023fatezero}.
We evaluate our \workname with two different spatial conditions: canny edge maps~\cite{canny1986computational} and depth maps~\cite{midas_ranftl2020towards}. A comparison of these controls can be found in Section~\ref{sec:ablation_edge_depth}.

\paragraph{Evaluation} %
We select the 25 object-centric videos from the public DAVIS dataset \cite{davis_pont20172017}, covering humans, animals, \textit{etc}. We manually design 115 prompts for these videos, spanning from stylization to object swap.
Besides, we also collect 50 Shutterstock videos~\cite{ShutterstockVideo} with 200 designed prompts.
We conduct both qualitative (see Section~\ref{exp:qualitative_results}) and quantitative comparisons (see Section~\ref{exp:quantitative_results}) with state-of-the-art methods including Rerender~\cite{rerender_yang2023rerender}, CoDeF~\cite{codef_ouyang2023codef} and TokenFlow~\cite{geyer2023tokenflow}.
We use their official codes with the default settings.

\subsection{Qualitative results}
\label{exp:qualitative_results}

In Figure~\ref{fig:main_comparsion}, we qualitatively compare our method with several representative approaches. 
Starting with a per-frame baseline directly applying I2I models, ControlNet, to each frame. 
Despite using a fixed random seed, this baseline often results in noticeable flickering, such as in the man's clothing and the tiger's fur.
CoDeF~\cite{codef_ouyang2023codef} produces outputs with significant blurriness when motion is big in input video, evident in areas like the man's hands and the tiger's face.
Rerender~\cite{rerender_yang2023rerender} often fails to capture large motions, such as the movement of paddles in the left example. Also, the color of the edited tiger's legs tends to blend in with the background.
TokenFlow~\cite{geyer2023tokenflow} occasionally struggles to follow the prompt, such as transforming the man into a pirate in the left example. 
It also erroneously depicts the tiger with two legs for the first frame in the right example, leading to flickering in the output video.
In contrast, our method stands out in terms of editing capabilities and overall video quality, demonstrating superior performance over these methods.
We highly encourage readers to refer to more video comparisons in our supplementary videos.

\subsection{Quantitative results}
\label{exp:quantitative_results}

\begin{table}[t]
\centering
\caption{\textbf{Quantitative comparison with existing V2V models.} The preference rate indicates the frequency the method is preferred among all the four methods in human evaluation. Runtime shows the time to synthesize a 4-second video with 512$\times$512 resolution on one A-100-80GB. Cost is normalized with our method.}
\small

\begin{tabular}{l|c|cc}
\toprule
\multirow{2}{*}{} & Preference rate & Runtime & \multirow{2}{*}{Cost $\downarrow$}\\
 & (mean $\pm$ std \%) $\uparrow$  & (mins) $\downarrow$  &   \\
\midrule
TokenFlow & 40.4 $\pm$ 5.3 & 15.8 & 10.5 $\times$ \\
Rerender & 10.2 $\pm$ 7.1 & 10.8 & 7.2 $\times$ \\
CoDeF & 3.5 $\pm$ 1.9 & 4.6 & 3.1 $\times$ \\
\workname(Ours) & \textbf{45.7} $\pm$ 6.4 & \textbf{1.5} & \textbf{1.0 $\times$} \\
\bottomrule
\end{tabular}

\label{tab:user_study}
\end{table}

\paragraph{User study}
We conducted a human evaluation to compare our method with three notable works: CoDeF~\cite{codef_ouyang2023codef}, Rerender~\cite{rerender_yang2023rerender}, and TokenFlow~\cite{geyer2023tokenflow}. The user study involves 25 DAVIS videos and 115 manually designed prompts. Participants are shown four videos and asked to identify which one has the best quality, considering both temporal consistency and text alignment.
The results, including the average preference rate and standard deviation from five participants for all methods, are detailed in Table~\ref{tab:user_study}. Our method achieved a preference rate of 45.7\%, outperforming CoDeF (3.5\%), Rerender (10.2\%), and TokenFlow (40.4\%).
During the evaluation, we observed that CoDeF struggles with significant motion in videos. The blurry constructed canonical images would always lead to unsatisfactory results. Rerender occasionally experiences color shifts and bright flickering. TokenFlow sometimes fails to sufficiently alter the video according to the prompt, resulting in an output similar to the original video.

\paragraph{Pipeline runtime}
We also compare runtime efficiency with existing methods in Table~\ref{tab:user_study}. Video lengths can vary, resulting in different processing times. Here, we use a video containing 120 frames (4 seconds video with FPS of 30). The resolution is set to 512 $\times$ 512.  Both our \workname model and Rerender~\cite{rerender_yang2023rerender} use a key frame interval of 4. We generate 31 keyframes by applying autoregressive evaluation twice, followed by RIFE~\cite{huang2022rife} for interpolating the non-key frames.
The total runtime, including image processing, model operation, and frame interpolation, is approximately 1.5 minutes. This is significantly faster than CoDeF (4.6 minutes), Rerender (10.8 minutes) and TokenFlow (15.8 minutes), being 3.1$\times$, 7.2$\times$, and 10.5 $\times$ faster, respectively.
CoDeF requires per-video optimization to construct the canonical image.
While Rerender adopts a sequential method, generating each frame one after the other, our model utilizes batch processing, allowing for more efficient handling of multiple frames simultaneously.
In the case of TokenFlow, it requires a large number of DDIM inversion steps (typically around 500) for all frames to obtain the inverted latent, which is a resource-intensive process.
We further report the runtime breakdown (Figure~\ref{fig:runtime_breakdown}) in the Appendix.

\begin{figure}[t]
  \begin{subfigure}{0.5\textwidth}
    \centering
    \includegraphics[width=\linewidth]{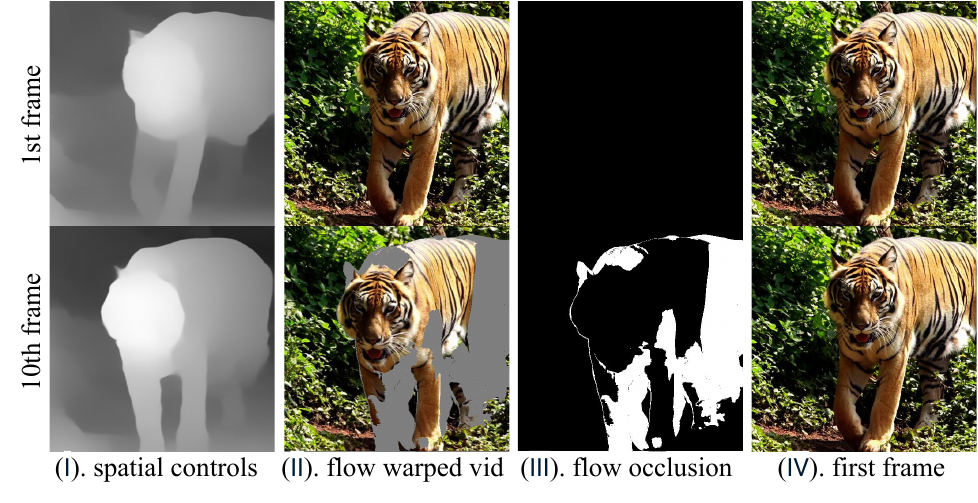}
    \vspace{-0.6cm}
    \caption{Condition types.}
    \vspace{0.3cm}
  \end{subfigure}
    \begin{subtable}{0.5\textwidth}
    \small
    \centering
        \begin{tabular}{cccccccc}
        \toprule
        \multicolumn{4}{c}{Condition choices}
         &  \multirow{2}{*}{Winning rate $\uparrow$} \\
         \cmidrule(lr){1-4}
         (I) & (II) & (III) & (IV) &  \\
        \midrule
          \checkmark &$\times$&$\times$&$\times$& 9\% \\
          \checkmark & \checkmark &$\times$&$\times$& 38\% \\
          \checkmark & \checkmark & \checkmark &$\times$  & 42 \% \\
        \bottomrule
        \end{tabular}
    \caption{Winning rate over our \workname (I + II + III + IV).}
  \end{subtable}
  
  \caption{\textbf{Ablation study of condition combinations.} (a) Four types of conditions. (b) The different combinations all underperform our final setting which combines all four conditions.}
  \vspace{-1em}
 \label{fig:condition_types}
\end{figure}

\subsection{Ablation study}

\paragraph{Condition combinations} 
\label{sec:ablation_conditions}
We study the four types of conditions in Figure~\ref{fig:condition_types}(a): (I) Spatial controls: such as depth maps~\cite{midas_ranftl2020towards}. (II) Flow warped video: frames warped from the first frame using optical flow. (III) Flow occlusion: masks indicate which parts are occluded (marked as white). (IV) First frame. 
We evaluate combinations of these conditions in Figure~\ref{fig:condition_types}(b), assessing their effectiveness by their winning rate against our full model which contains all four conditions.
The spatial-only condition achieved a 9\% winning rate, limited by its lack of temporal information. Including flow warped video significantly improved the winning rate to 38\%, underscoring the importance of temporal guidance.
We use gray pixels to indicate occluded areas, which might blend in with the original gray colors in the images. 
To avoid potential confusion, we further include a binary flow occlusion mask, which better helps the model to tell which part is occluded or not.
The winning rate is further improved to 42\%.
Finally, we added the first frame condition to provide better texture guidance, particularly useful when the occlusion mask is large and few original pixels remain.

\begin{figure}[t]
    \centering
	\includegraphics[width=1.0\columnwidth]{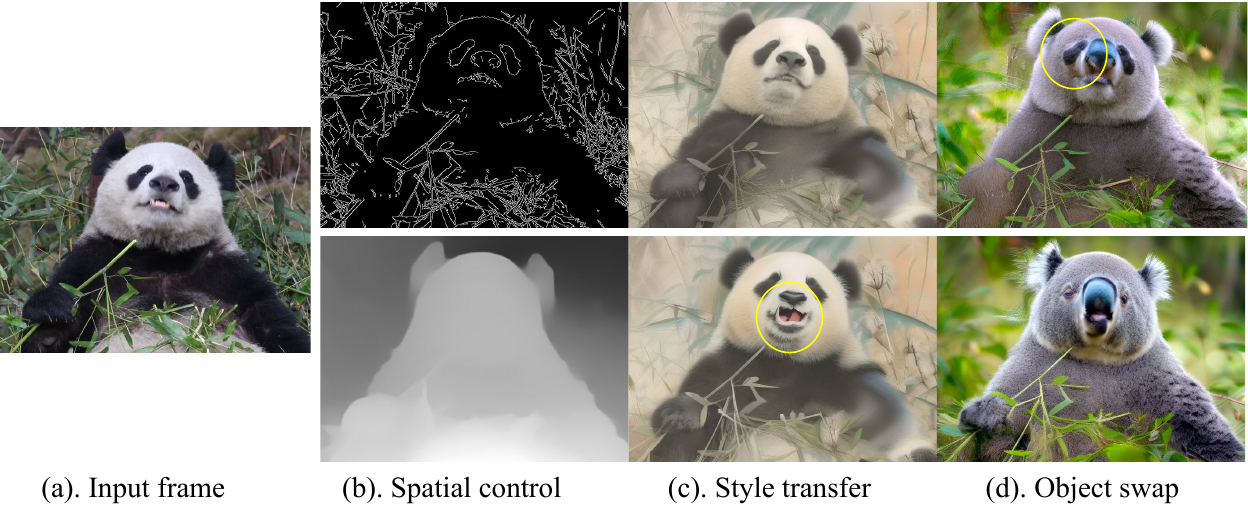}
	\caption{\textbf{Ablation study of different spatial conditions.} Canny edge and depth map are estimated from the input frame. Canny edge provides more detailed controls (good for stylization) while depth map provides more editing flexibility (good for object swap).}
	\label{fig:canny_depth}
\end{figure}

\begin{figure}[t]
    \centering
	\includegraphics[width=0.8\columnwidth]{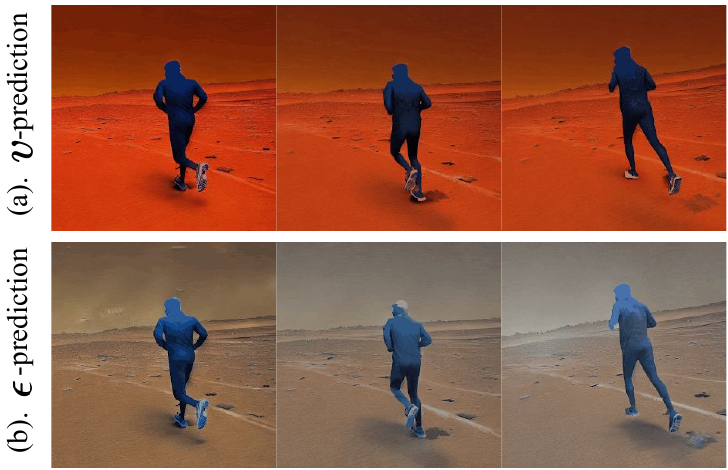}
	\caption{\textbf{Ablation study of different parameterizations.} $\epsilon$-prediction often predicts unnatural global color while $v$-prediction doesn't. Prompt: \texttt{'a man is running on Mars'}. 
 }
	\label{fig:v_epsilon}
\end{figure}

\paragraph{Different control type: edge and depth}
\label{sec:ablation_edge_depth}
We study two types of spatial conditions in our \workname: canny edge~\cite{canny1986computational} and depth map~\cite{midas_ranftl2020towards}. Given an input frame as shown in Figure~\ref{fig:canny_depth}(a), the canny edge retains more details than the depth map, as seen from the eyes and mouth of the panda. 
The strength of spatial control would, in turn, affect the video editing. 
For style transfer prompt \texttt{'A Chinese ink painting of a panda eating bamboo'}, as shown in Figure~\ref{fig:canny_depth}(c), the output of canny condition could keep the mouth of the panda in the right position while the depth condition would guess where the mouth is and result in an open mouth.
The flexibility of the depth map, however, would be beneficial if we are doing object swap with prompt \texttt{'A koala eating bamboo'}, as shown in Figure~\ref{fig:canny_depth}(d); the canny edge would put a pair of panda eyes on the face of the koala due to the strong control, while depth map would result in a better koala edit.
During our evaluation, we found canny edge works better when we want to keep the structure of the input video as much as possible, such as stylization. The depth map works better if we have a larger scene change, such as an object swap, which requires more considerable editing flexibility.

\paragraph{$v$-prediction and $\epsilon$-prediction}
\label{sec:v_epsilon}
While $\epsilon$-prediction is commonly used for parameterization in diffusion models, we found it may suffer from unnatural global color shifts across frames, as shown in Figure~\ref{fig:v_epsilon}. Even though all these two methods use the same flow warped video, the $\epsilon$-prediction introduces an unnatural grayer color. 
This phenomenon is also found in Imagen-Video~\cite{imagenvideo_ho2022imagen}.

\subsection{Limitations} 
\begin{figure}[t]
    \centering
	\includegraphics[width=1.0\columnwidth]{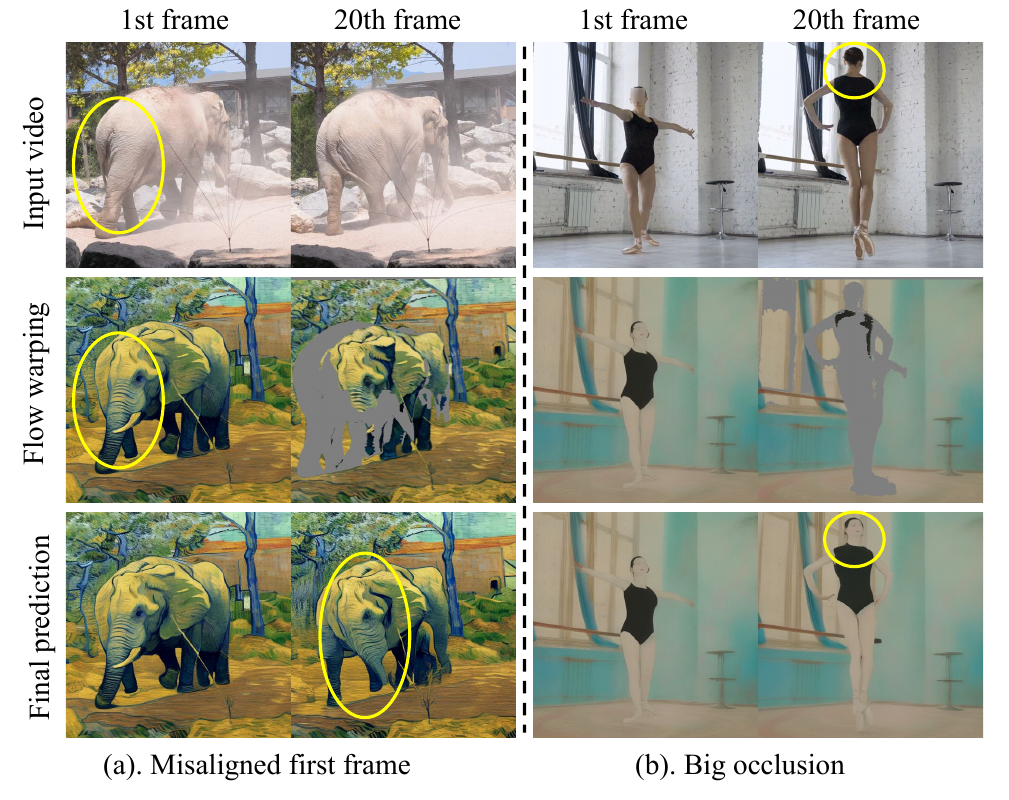}
	\caption{\textbf{Limitations of \workname.} Failure cases include (a) the edited first frame doesn't align structurally with the original first frame, and (b) large occlusions caused by fast motion.}
	\label{fig:limitations}
 \vspace{-1em}
\end{figure}

Although our \workname achieves significant performance, it does have some limitations. First, our \workname heavily relies on the first frame generation, which should be structurally aligned with the input frame. As shown in Figure~\ref{fig:limitations}(a), the edited first frame identifies the hind legs of the elephant as the front nose. The erroneous nose would propagate to the following frame and result in an unsatisfactory final prediction. The other challenge is when the camera or the object moves so fast that large occlusions occur. In this case, our model would guess, sometimes hallucinate, the missing blank regions. As shown in Figure~\ref{fig:limitations}(b), when the ballerina turns her body and head, the entire body part is masked out. Our model manages to handle the clothes but turns the back of the head into the front face, which would be confusing if displayed in a video. 

\section{Conclusion}
\label{sec:conclusion}

In this paper, we propose a consistent video-to-video synthesis method using joint spatial-temporal conditions. 
In contrast to prior methods that strictly adhere to optical flow, our approach incorporates flow as a supplementary reference in synergy with spatial conditions. 
Our model can adapt existing image-to-image models to edit the first frame and propagate the edits to consecutive frames. Our model is also able to generate lengthy videos via autoregressive evaluation.
Both qualitative and quantitative comparisons with current methods highlight the efficiency and high quality of our proposed techniques.

\section{Acknowledgments}

We would like to express sincere gratitude to Yurong Jiang, Chenyang Qi, Zhixing Zhang, Haoyu Ma, Yuchao Gu, Jonas Schult, Hung-Yueh Chiang, Tanvir Mahmud, Richard Yuan for the constructive discussions. 

Feng Liang and Diana Marculescu were supported in part by the ONR Minerva program, iMAGiNE - the Intelligent Machine Engineering Consortium at UT Austin, and a UT Cockrell School of Engineering Doctoral Fellowship.

{\small
\bibliographystyle{ieeenat_fullname}
\bibliography{sections/11_references}
}

\ifarxiv \appendix \section{Webpage Demo}
We highly recommend looking at our demo web by opening the \url{https://jeff-liangf.github.io/projects/flowvid/} to check the video results.

\section{Quantitative comparisons }

\subsection{CLIP scores}
Inspired by previous research, we utilize CLIP \cite{clip_radford2021learning} to evaluate the generated videos' quality. Specifically, we measure 1) Temporal Consistency (abbreviated as Tem-Con), which is the mean cosine similarity across all sequential frame pairs, and 2) Prompt Alignment (abbreviated as Pro-Ali), which calculates the mean cosine similarity between a given text prompt and all frames in a video. Our evaluation, detailed in Table~\ref{tab:comparison}, includes an analysis of 116 video-prompt pairs from the DAVIS dataset. Notably, CoDeF~\cite{codef_ouyang2023codef} and Rerender~\cite{rerender_yang2023rerender} exhibit lower scores in both temporal consistency and prompt alignment, aligning with the findings from our user study. Interestingly, TokenFlow shows superior performance in maintaining temporal consistency. However, it is important to note that TokenFlow occasionally underperforms in modifying the video, leading to outputs that closely resemble the original input. Our approach not only secures the highest ranking in prompt alignment but also performs commendably in temporal consistency, achieving second place.

\begin{table}[h]
    \label{tab:clip_score}
    \caption{\textbf{CLIP score comparisons}. 'Tem-Con' stands for temporal consistency, and 'Pro-Ali' stands for prompt alignment.}
    \label{tab:comparison}
    \small
    \centering
    \begin{tabular}{l|cc}
        \toprule
        Method & Tem-Con $\uparrow$ & Pro-Ali $\uparrow$\\
        \midrule 
        CoDeF~\cite{codef_ouyang2023codef} & 96.98 & 30.83  \\
        Rerender~\cite{rerender_yang2023rerender} & 96.88 & 31.84  \\
        TokenFlow~\cite{geyer2023tokenflow} & \textbf{97.30} & 33.11 \\
        Ours & 97.08 & \textbf{33.20}\\
    \bottomrule
    \end{tabular}
\end{table}

\subsection{Runtime breakdown}

We benchmark the runtime with a 512 $\times$ 512 resolution video containing 120 frames (4 seconds video with FPS of 30). 
Our runtime evaluation was conducted on a 512 × 512 resolution video comprising 120 frames, equating to a 4-second clip at 30 frames per second (FPS). Both our methods, FlowVid, and Rerender~\cite{rerender_yang2023rerender}, initially create key frames followed by the interpolation of non-key frames. For these techniques, we opted for a keyframe interval of 4. FlowVid demonstrates a marked efficiency in keyframe generation, completing 31 keyframes in just 1.1 minutes, compared to Rerender's 5.2 minutes. This significant time disparity is attributed to our batch processing approach in FlowVid, which handles 16 images simultaneously, unlike Rerender's sequential, single-image processing method.
In the aspect of frame interpolation, Rerender employs a reference-based EbSynth method, which relies on input video's non-key frames for interpolation guidance. This process is notably time-consuming, requiring 5.6 minutes to interpolate 90 non-key frames. In contrast, our method utilizes a non-reference-based approach, RIFE~\cite{huang2022rife}, which significantly accelerates the process.
Two other methods are CoDeF~\cite{codef_ouyang2023codef} and TokenFlow~\cite{geyer2023tokenflow}, both of which necessitate per-video preparation. Specifically, CoDeF involves training a model for reconstructing the canonical image, while TokenFlow requires a 500-step DDIM inversion process to acquire the latent representation. CoDeF and TokenFlow require approximately 3.5 minutes and 10.4 minutes, respectively, for this initial preparation phase.

\begin{figure}[t]
    \centering
	\includegraphics[width=1.0\columnwidth]{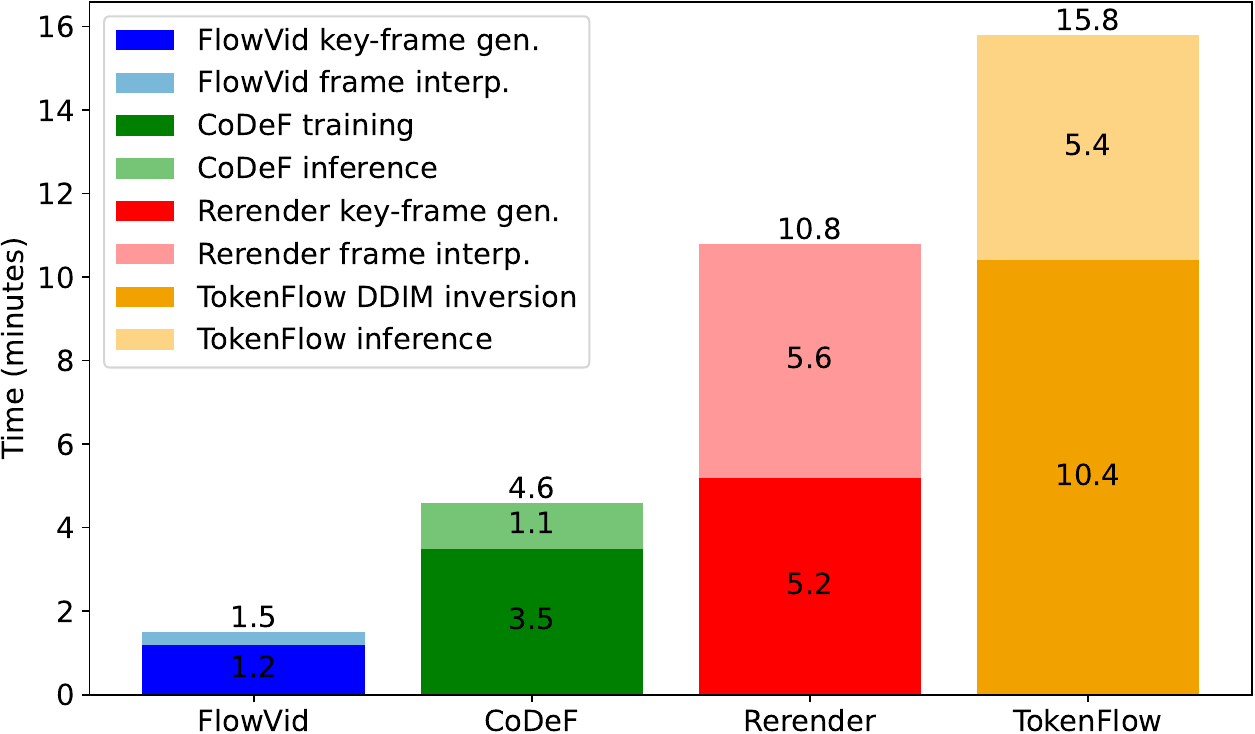}
	\caption{\textbf{Runtime breakdown} of generating a 4-second 512 $\times$ 512 resolution video with 30 FPS. Time is measured on one A100-80GB GPU.}
	\label{fig:runtime_breakdown}
\end{figure}
 \fi

\end{document}